%% file: 0-paper_lpl.tex
\documentclass[letterpaper, 10 pt, conference]{IEEEtran}  % Comment this line out if you need a4paper

\IEEEoverridecommandlockouts                              % This command is only needed if 
\usepackage{amsmath,amsfonts}
\usepackage{algorithm}
\usepackage{array}
\usepackage[caption=false,font=normalsize,labelfont=sf,textfont=sf]{subfig}
\usepackage{textcomp}
\usepackage{stfloats}
\usepackage{url}
\usepackage{verbatim}
\usepackage{graphicx}
\usepackage{cite}
\usepackage{xspace}
\usepackage{multirow} 
\usepackage{color}
\usepackage{makecell}
\usepackage{graphicx}
\usepackage{float}
\usepackage{textcomp,booktabs}

\usepackage{algorithm}
\usepackage{algpseudocode}
\usepackage{amsmath}

\usepackage{colortbl}
\usepackage{bbding}
\usepackage{CJKutf8}
\usepackage{colortbl}
\usepackage{xcolor}

\definecolor{gray_loc}{RGB}{118,113,113}
\definecolor{bule_lane}{RGB}{91,155,213}
\definecolor{green_rule}{RGB}{116,171,79}
\definecolor{orange_plan}{RGB}{244,173,124}
\definecolor{yellow_drive}{RGB}{255,192,0}

\definecolor{mygray}{gray}{.9}
\definecolor{pastelyellow}{rgb}{1.0, 0.902, 0.59}
\definecolor{myyellow}{rgb}{1.0, 0.988, 0.9}
\newcommand{\systemname}{{LFP}\xspace}
\begin{document}

\title{\LARGE \bf
\systemname: Efficient and Accurate End-to-End Lane-Level Planning via Camera-LiDAR Fusion
}
\author{Guoliang You$^{1}$, Xiaomeng Chu$^{1}$, Yifan Duan$^{1}$, Xingchen Li$^{1,3}$, Sha Zhang$^{1,4}$, \\Jianmin Ji$^{1}$, Yanyong Zhang$^{2}$,~\IEEEmembership{Fellow,~IEEE}
\thanks{$^1$ School of Computer Science and Technology, University of Science and Technology of China (USTC), Hefei 230026, China}
\thanks{$^2$ School of Artificial Intelligence and Data Science, University of Science and Technology of China (USTC), Hefei 230026, China}
\thanks{$^3$ SenseTime Research, Shanghai 200000, China}
\thanks{$^4$ Shanghai AI Laboratory, Shanghai 200000, China}
\thanks{ $^\dag$ Corresponding author. {\tt\small yanyongz, jianmin@ustc.edu.cn}}}

\maketitle
\thispagestyle{empty}
\pagestyle{empty}

%%%%%%%%%%%%%%%%%%%%%%%%%%%%%%%%%%%%%%%%%%%%%%%%%%%%%%%%%%%%%%%%%%%%%%%%%%%%%%%%
\begin{abstract}
Multi-modal systems enhance performance in autonomous driving but face inefficiencies due to indiscriminate processing within each modality. 
Additionally, the independent feature learning of each modality lacks interaction, which results in extracted features that do not possess the complementary characteristics.
These issue increases the cost of fusing redundant information across modalities.
To address these challenges, we propose targeting driving-relevant lane elements, which effectively reduces the volume of LiDAR features while preserving critical information. This approach enhances interaction at the lane level between the image and LiDAR branches, allowing for the extraction and fusion of their respective advantageous features.
Building upon the camera-only framework PHP\cite{php}, 
we introduce the \underline{L}ane-level camera-LiDAR \underline{F}usion \underline{P}lanning (\systemname) method, which balances efficiency with performance by using lanes as the unit for sensor fusion.
Specifically, we design three novel modules to enhance efficiency and performance.
For efficiency, we propose an image-guided coarse lane prior generation module that forecasts the region of interest (ROI) for lanes and assigns a confidence score, guiding LiDAR processing and fusion. The LiDAR feature extraction modules leverages lane-aware priors from the image branch to guide sparse sampling for pillar features, retaining essential features. 
For performance, the lane-level cross-modal query integration and feature enhancement module uses confidence score from ROI to combine low-confidence image queries with LiDAR queries, extracting complementary depth features. These features then enhance the corresponding low-confidence image features, compensating for the lack of depth. 
Experiments on the Carla benchmarks show that our method achieves state-of-the-art performance in both driving score and infraction score, with maximum improvement of 15\% and 14\% over existing algorithms, respectively, while maintaining high frame rate of 19.27 FPS.
\end{abstract}

%%%%%%%%%%%%%%%%%%%%%%%%%%%%%%%%%%%%%%%%%%%%%%%%%%%%%%%%%%%%%%%%%%%%%%%%%%%%%%%%

%%%%%%%%% BODY TEXT
\input{2-introduction}
\input{3-related_work}
\input{4-methodology}
\input{5-experiments}
\input{6-conclusions}

\bibliographystyle{IEEEtran}
\bibliography{IEEEabrv, 7-reference}

\end{document}

%% file: 2-introduction.tex
\section{INTRODUCTION}
\begin{figure}[t]
  \centering
  \includegraphics[width=\columnwidth]{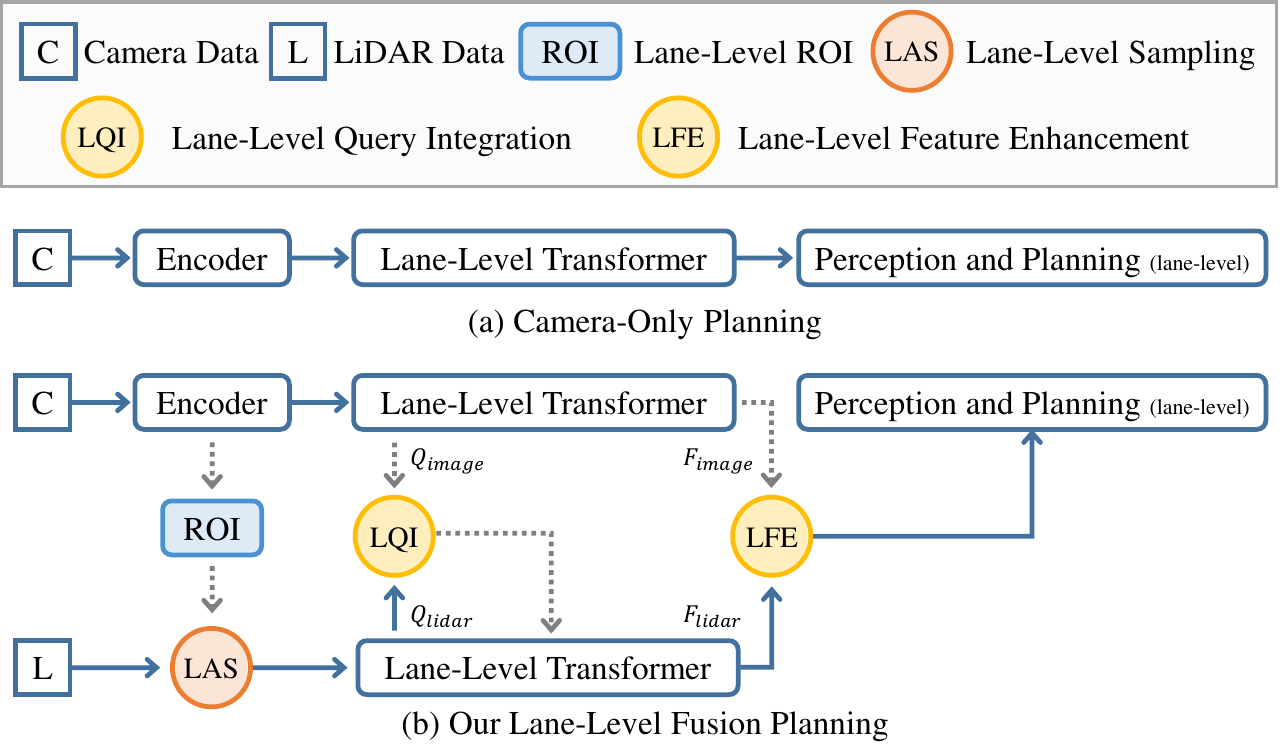}
  \caption{An illustration comparing (a) the Camera-Only end-to-end planning with (b) our proposed lane-level Camera-LiDAR fusion end-to-end planning, where we utilize the geometric lane priors from images to guide the LiDAR branch in efficiently extracting depth features that the image branch lacks.}
  \label{fig:idea}
\end{figure}

\IEEEPARstart{A}{utonomous} driving plays a vital role in improving the efficiency and safety of transportation systems~\cite{autonomous-survey}, intersecting the fields of computer vision and robotics~\cite{perception-3D-survey1,calib-survey, slam-survey}. A key challenge lies in devising efficient and accurate planning~\cite{planning-survey}.
Autonomous driving planning algorithms can be broadly categorized into two types: rule-based planning~\cite{perception-3d-BEVFormer,lane-bev-laneDet, rule-planning-rrt, rule-planning-RRV} and end-to-end learning-based planning~\cite{e2e-planning-survey, e2e-nvidia,e2e-pathrl,m2m-stp3, m2m-uniad-Goal-oriented,m2m-neat}.
Traditional rule-based planning leverages heavy algorithms~\cite{lane-bev-laneDet, perception-3d-BEVFormer, perception-3d-rayformer} to precisely detect lane and position vehicles and pedestrians. Multi-modal algorithms~\cite{multi-sensor-bevfusion, multi-sensor-transfusion} are also employed to augment perception capabilities. These are typically followed by optimization techniques like the rapidly exploring random tree (RRT)~\cite{rule-planning-rrt} to identify safe paths, with high computational cost. 
End-to-end learning-based planning algorithms, such as ST-P3~\cite{m2m-stp3} and UniAD~\cite{m2m-uniad-Goal-oriented}, leverage camera data and deep learning to streamline the planning process, reducing computational overhead~\cite{e2e-planning-survey, e2e-nvidia, e2e-pathrl}. 
Despite this, pure vision systems often struggle with depth perception, which impairs the accuracy of environmental understanding and, by extension, the efficacy of planning algorithms.
To overcome these, methods such as TransFuser~\cite{m2m-transfuser} and Interfuser~\cite{m2m-interfuser} incorporate LiDAR's depth data into the end-to-end planning framework through attention mechanisms, significantly boosting perception and planning accuracy. 
For better fusion, the bird's eye view (BEV) offers an advantageous perspective for multi-modal fusion, providing a holistic view that streamlines the integration of camera and LiDAR data, as demonstrated by Think Twice~\cite{multi-modal-thinktwice}.
However, indiscriminate processing of all BEV information results in inefficiency, as not all environmental elements are critical to planning. Moreover, the lack of interactive feature extraction between different modalities does not fully exploit their complementary strengths.

Given these challenges, the question arises: Can we achieve efficient and complementary feature extraction and fusion between LiDAR and camera? 
In response, we introduce \systemname, a novel lane-level camera-LiDAR fusion method that extends PHP~\cite{php} for multi-modal fusion. Using lane priors from the camera branch, \systemname minimizes essential LiDAR features, retaining only key information while improving the interaction between LiDAR and camera data, thus ensuring the extraction of complementary features.

Specifically, in \systemname, the image branch, termed the image-guided coarse lane prior generation, processes images from multiple cameras to distill features. These features are submitted to a transformer that extracts lane-level image features. In parallel, the coarse lane detection module leverages the semantic richness of images to delineate lane geometric priors and quantify their confidence.
The LiDAR branch, termed lane-level LiDAR feature extraction, leverages the Lane ROI for sparse sampling of LiDAR pillars, focusing critical lane regions. 
The fusion branch, guided by lane priors and termed lane-level cross-modal query integration and feature enhancement, first integrates the lane-level queries from both the image and LiDAR branches, refining the LiDAR branch's transformer retrieval of lane-level features. Then, through a weighted integration strategy, it maximizes the strengths of each modality, thereby enhancing the system's overall perception and planning capabilities.
Finally, the fused features are utilized in the lane-level planning module to produce accurate perception and planning results.

This work shows that a lane-level camera-LiDAR fusion method, guided by lane priors from images, efficiently targets key driving elements within the LiDAR point cloud, avoiding unnecessary computations. Meanwhile, it promotes lane-level interaction to extract and fuse the most advantageous features from both the camera and LiDAR branches, enhancing the precision of planning.

In summary, our main contributions are as follows:
\begin{itemize}
\item We introduce \systemname, a lane-level camera-LiDAR fusion planning algorithm that efficiently integrates LiDAR, enhancing feature depth and semantic richness via camera-LiDAR interaction, boosting planning performance.
\item We developed a method with a coarse prior extraction algorithm that generates lane-level priors, enabling sparse LiDAR feature sampling and guiding the extraction and fusion of complementary features between camera and LiDAR branches at both the query and feature levels.
\item We conducted experiments on the Carla benchmarks, demonstrating that \systemname achieved state-of-the-art performance in both driving score and infraction score, with maximum improvement of 15\% and 14\%, and maintained high computational efficiency at up to 19.27 FPS.
\end{itemize}

%% file: 3-related_work.tex
\section{RELATED WORK}
\subsection{Vision-Based End-to-End Autonomous Driving}
Vision-based End-to-end learning methods, such as those presented by Amini et al.~\cite{vision-based-mit-variational} and Bojarski et al.~\cite{vision-based-nvidia}, have made strides in mapping raw visual inputs directly to policies.
In addition, researchers have explored various strategies. Codevilla et al.~\cite{vision-based-cilrs} and Chen et al.~\cite{vision-based-lbc} have delved into the nuances of behavior cloning, identifying its limitations, and proposing novel techniques to enhance the learning process. The incorporation of reinforcement learning, as shown by Zhang et al.~\cite{vision-based-roach}, has further enriched training regimens.
Moreover, the advent of spatial-temporal feature learning, as demonstrated by Hu et al.~\cite{vision-based-stp3}, and the innovative vectorized scene representation by Jiang et al.~\cite{vision-based-vad}, has paved the way for more effective planning. 
For comprehensive driving understanding, these methods may still underperform when relying solely on visual data. Systems that integrate additional sensors such as LiDAR, which provide critical depth information, generally perform better.

\subsection{Multi-Modal End-to-End Autonomous Driving}
The advancement of autonomous driving has been significantly propelled by the integration of multi-modal data. 
Recognizing the value of additional information, Chitta et al.~\cite{multi-modal-transfuser} and Chen et al.~\cite{multi-modal-m2m-lav} leveraged camera and LiDAR data to achieve multi-modal fusion, showcasing enhanced planning performance. 
By emphasizing multi-modal temporal and global reasoning in driving scenarios, as showcased in their previous work~\cite{multi-modal-shao2023reasonnet}, Shao et al.~\cite{multi-modal-shao2024lmdrive} have further demonstrated the capability of using language instructions and multi-modal sensor data as input to generate control signals for driving.
Additional, interpretability and safety in autonomous systems have been addressed by Shao et al.~\cite{multi-modal-interfuser} through the development of transformer-based sensor fusion models. Jia et al. ~\cite{multi-modal-driveadapter} proposed a novel multi-modal method that decouples perception and planning, allowing for more flexible system design.
Furthermore, Jia et al. proposed ThinkTwice~\cite{multi-modal-thinktwice}, which focused on developing scalable decoders for multi-modal end-to-end planning, and DriveMLM by Wang et al.~\cite{multi-modal-drivemlm}, which aligned multi-modal large language models with behavioral planning states.
Collectively, these works show how multi-modal data fusion benefits autonomous driving.

%% file: 4-methodology.tex
\section{METHODOLOGY}
\begin{figure*}[t]
  \centering
  \includegraphics[width=0.95\linewidth]{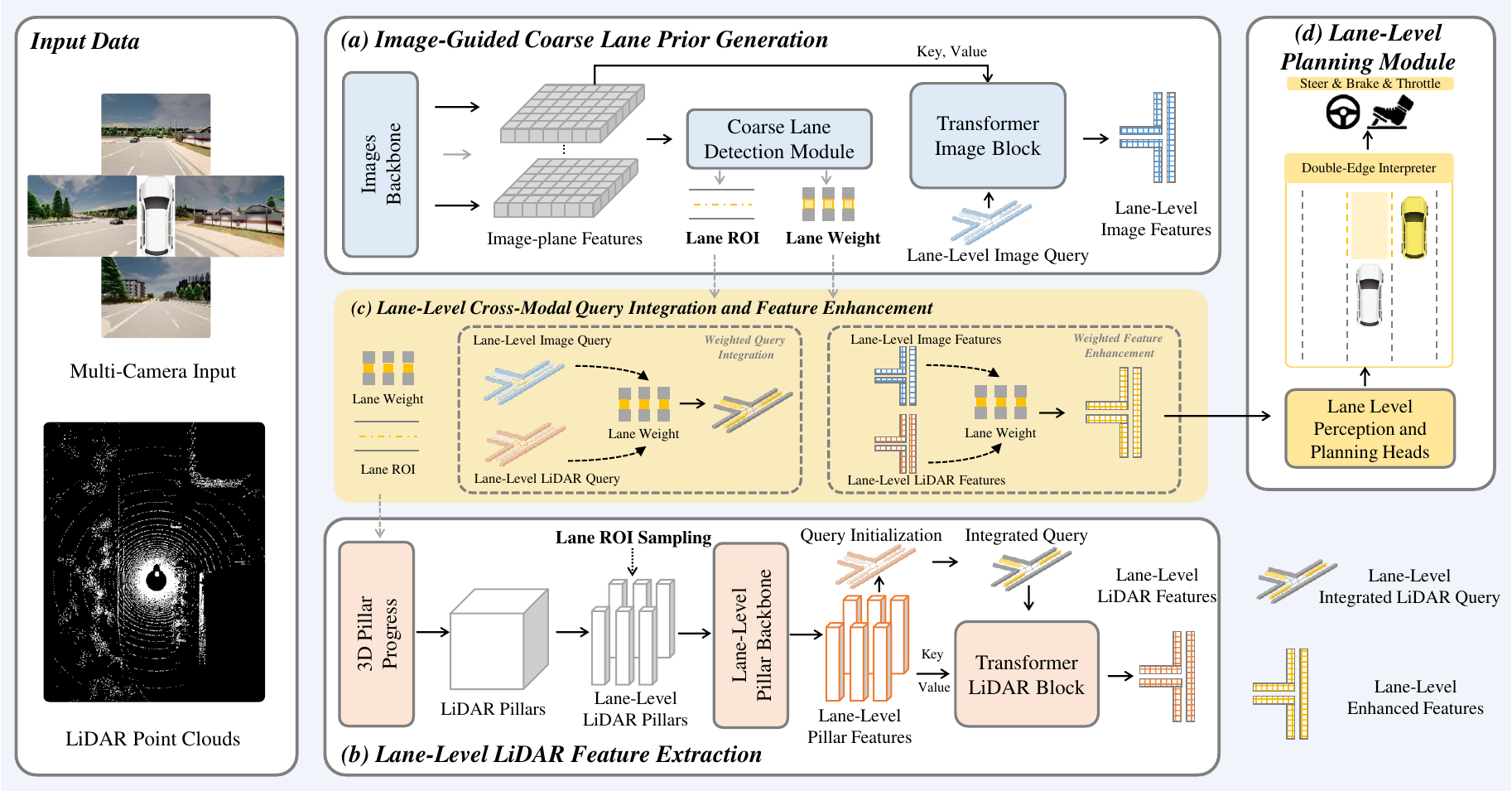}
  \caption{The \systemname integrates image and LiDAR through four modules: (a) The image-guided coarse lane prior generation module, which extracts lane-level image features and generates coarse lane priors (Lane ROI and Lane Weight); (b) The lane-level LiDAR feature extraction module, which performs pillar-based sampling guided by lane priors to focus on lane areas and extracts lane-level LiDAR features; (c) The lane-level cross-modal query integration and feature enhancement module, which integrates queries and features from both image and LiDAR at the query and feature levels; (d) The lane-level planning module, which processes the lane-level enhanced features, outputs lane-level perception and planning results, and converts them into vehicle control signals.}
  \label{fig:lpl_arc}
\end{figure*}

\subsection{Preliminary}
\label{method:dual-lane-data}
In \systemname, we continue to employ the double-edge data structure from the PHP framework\cite{php}, focusing on lane-level planning.
PHP defines $N_{d}$ double-edge data ($l^{i}_{d}$) that incorporate lane-level and point-level traffic information data to describe the environment $L$, where $L=\left\{l_d^i\right\}_{i=0}^{\mathrm{N}_d}$,
\begin{equation}
\begin{aligned}
 { l }_{ {d }}^i &=\left( { edge  }_{ {l }}^i,  { edge  }_{ {r }}^i,  { int }_i,  { dir }_i\right),\\
 { int }_i &=\{0 \ { or\ } 1\},  { dir }_i=\{0 \ { or\ } 1\}.
 \end{aligned}
\end{equation}
Here, $l^{i}_{d}$ includes the lane's left and right $edge_i$, along with its lane-level intersection $int_i$ and direction $dir_i$ attributes.
Within the double-edge, $int_{i}$ and $dir_{i}$ indicate whether a lane is an intersection and if the lane's direction is aligned with the direction of the ego vehicle's travel respectively. 
$edge^i$ encompasses 3D point and attributes for corresponding points, including occupancy and planning attributes that indicate the point level. Each $edge$ consists of $\frac{N_p}{2}$ elements, defined as:
\begin{equation}
\begin{aligned}
 {edge}^{i} &=\left\{ { point }^{j}, { occ }^{j}, { plan }^{j}\right\}_{j=0}^{\frac{N_p}{2}},\\
 {point}^{j} &=\{\mathrm{x}, \mathrm{y}, \mathrm{z}\}, \operatorname{occ}^{j}=\{0  {\ or\ } 1\},  { plan }^{j}=\{0  {\ or\ } 1\},
\end{aligned}
\end{equation}
where $occ^{j}$ and $plan^{j}$ indicate whether the lane is occupied by agents (e.g., pedestrians, vehicles) and whether it is selected for planning. 
Ultimately, we utilize the lane-level planning module (Sec\ref{lane-level-planning}) to predict this structure, enabling the output of lane-level perception and planning results.

\subsection{Image-Guided Coarse Lane Prior Generation}
\label{method:img-guide-lane-prioritization}
This module extracts coarse lane priors from multi-view images, while using lane-level image queries to extracts lane-level features from the environment, as depicted in Figure \ref{fig:lpl_arc}(a).

Initially, for each image input $\mathbf{I} \in \mathbb{R}^{3 \times H’ \times W’}$ , we employ a ResNet-50~\cite{tool-ResNet} to extracting features $\mathbf{f} \in \mathbb{R}^{C \times H \times W}$. The values for $C$, $H$, and $W$ are defined as $C=256$, $H = \frac{H’}{32}$, and $W = \frac{W’}{32}$. The dimension of the transformer hidden layer is $E$. For each feature ${\mathbf{f}}$, we applies a ${1×1}$ convolution to generate a lower-channel feature $\mathbf{z} \in \mathbb{R}^{E \times H \times W}$. Next, we simplify the spatial dimensions of ${\mathbf{z}}$ into a sequence, forming ${E \times HW}$ tokens. A fixed sinusoidal positional encoding $\mathbf{e} \in \mathbb{R}^{E \times HW}$ is then added to each token to preserve positional information within each sensor input: $\mathbf{v}_i^{(x, y)}=\mathbf{z}_i^{(x, y)}+\mathbf{e}^{(x, y)}$, where ${\mathbf{z}_i}$ represents the tokens extracted from the ${i-th}$ view, and ${x}$ and ${y}$ denote the token's coordinate index in that sensor.
Subsequently, for the coarse lane detection module processes the feature vectors \(\mathbf{v}_i\) to extract a set of \(N_d \times N_p \times 3\) dimensional coarse lane ROI (\(R_{lane}\)), which represent the rough 3D geometric information of the lane boundaries points ($point_j$) within the double-edge. Concurrently, it generates lane weights (\(W_{lane}\)) that quantify the confidence of these lane priors. This process is accomplished through a sequence of linear layers followed by ReLU activations.
Finally, we concatenate \(\mathbf{v}_i\) from all images and pass them through a transformer image block comprising ${K}$ standard transformer encoder and decoder layers. 
The transformer image block leverages \( N_d \) lane-level image queries \( \mathbf{q}_{lane-level} \in \mathbb{R}^{E \times N_p} \) to process and extract lane-level features \( \mathbf{f}_{image} \), focusing on driving-relevant areas.

\subsection{Lane-Level LiDAR Feature Extraction}
\label{method:lane-level-lidar-featrue}
This module efficiently extracts driving-relevant and lane-level features from LiDAR by leveraging image-based lane priors to focus on critical depth data, minimizing computational redundancy, as shown in Figure \ref{fig:lpl_arc}(b). 

The process begins with the pillar processing of the LiDAR points \( P \in \mathbb{R}^{N \times 3} \), where \( N \) represents the number of points. By processing the LiDAR point within pillars, where the height of the pillars is adjusted to match the actual height of the LiDAR points, we obtain a LiDAR pillar set \( V_{point} \).
However, direct feature encoding from these pillars still includes irrelevant data for driving tasks, such as trees and buildings. To retain the key depth information that the image branch lacks and reduce computational redundancy, we employ a lane-level sampling operation based on the image branch's lane ROI priors (\( R_{lane} \)). Utilizing the geometric positions of the lane ROI priors, we identify and retain the closest pillars to these priors, resulting in a set of lane-level pillars \( V_{lane} \)  with dimensions \( N_d \times N_p \times C\), where C  denotes the dimension of each pillars. 
This selective retention ensures that only the pillars critical to driving are preserved, allowing the LiDAR branch to focus on the most relevant data.

Subsequently, the lane-level pillar backbone processes the lane-level pillars \( V_{lane} \) to obtain the lane-level pillar features \( \mathbf{f}_{lane} \). These features are then used to initialize the lane-level LiDAR queries \( \mathbf{q}_{lidar} \) in the transformer LiDAR block, which have dimensions \( N_d \times N_p \times E \).
The transformer LiDAR block then leverages these queries to perform feature extraction through \( K \) layers of multi-head self-attention. The primary objective is to refine the lane-level pillar features into lane-level LiDAR features \( \mathbf{f}_{lidar} \). During this stage, we also prepare for the cross-modality query integration, which will be detailed in Sec\ref{method:lane-level-cross-modal}. This integration is anticipated to incorporate the image branch's query \( \mathbf{q}_{image} \) into the LiDAR query \( \mathbf{q}_{lidar} \), thereby enabling the LiDAR branch to target on the lane regions of interest as identified by the image branch.

\subsection{Lane-Level Cross-Modal Query Integration and Feature Enhancement}
\label{method:lane-level-cross-modal}
This section details the cross-modal query integration, harnessing image branch semantics to guide LiDAR depth feature extraction, complementing the image's depth limitations. It also outlines a feature enhancement strategy that merges branch strengths for efficient lane-level feature fusion, optimizing performance and computational efficiency, as in Figure~\ref{fig:lpl_arc}(c).
Specifically, the image branch, enriched with semantic information but limited in depth, excels at concentrating feature queries around the lanes. In contrast, the LiDAR branch, proficient in capturing depth information, may struggle to identify lane due to the lack of semantic cues. To address this, we propose a weighted integration of the image branch's query ${q}_{image}$ and the LiDAR branch's initialized query \( \mathbf{q}_{lidar} \), with weights determined by the lane weights \( \mathbf{w}_{lane} \) associated with the lane ROI. This integration results in a lane-level LiDAR integrated query, \( \mathbf{q}_{integrated} \), adept at querying the lane-level pillar features \( \mathbf{f}_{lane} \) to extract the critical depth features that the image lacks. 
The integration enhances the focus of LiDAR queries on areas where image depth cues are insufficient, directing the LiDAR branch to supplement the image depth features with precise depth information. \( \mathbf{q}_{integrated} \) is defined as:

\begin{equation}
\begin{aligned}
\mathbf{q}_{\text {integrated }}=(1-\alpha) \cdot \mathbf{q}_{\text { image}} + \alpha \cdot \mathbf{q}_{\text {lidar }},
\end{aligned}
\end{equation}
where \( \alpha \) represents the weight derived from the lane ROI.
After both branches have processed their sensor data and obtained lane-level features, a fusion of these features is necessary to leverage the semantic prowess of the image branch and the depth acuity of the LiDAR branch. We achieve this by employing a similar weighted fusion approach, combining the lane-level features from the image branch \( \mathbf{f}_{image} \) and the LiDAR branch \( \mathbf{f}_{lidar} \) with weights \( \mathbf{w}_{lane} \), leading to an enriched set of lane-level enhanced features \( \mathbf{f}_{enhanced} \):
\begin{equation}
\mathbf{f}_{\text {enhanced }}=\beta \cdot \mathbf{f}_{\text {image }}+(1-\beta) \cdot \mathbf{f}_{\text {lidar }},
\end{equation}
where \( \beta \) is the weight that balances contributions from both branches, derived from the lane ROI.

\subsection{Lane-Level Planning Module}
\label{lane-level-planning}
This module predicts lane-level perception and planning tasks using features $\mathbf{f}_{\text {enhanced }}$, including a lane-level perception and planning head for double-edge data structure and a double-edge interpreter for safe planning results, as in Figure~\ref{fig:lpl_arc}(d).

\noindent{\textbf{Lane-Level Perception and Planning Heads.}}
\systemname integrates a series of prediction heads for the double-edge. A regression head predicts boundary points \( point^j \), and four classification heads forecast attributes of intersection \( int_i \), direction \( dir_i \), occupancy \( occ^j \), and planning \( plan^j \) within the traffic scenario.

\noindent{\textbf{Double-Edge Interpreter.}}
The double-edge interpreter translating the rich geometric and attributive lane-level information from the double-edge into executable path by a controller.
Notably, by leveraging the image branch's strength in traffic signal recognition, we incorporate speed and traffic signal queries to predict velocity and traffic conditions. These predictions, including speed and path, are fed into the interpreter and then directly utilized by controllers~\cite{path_follow} to convert them into control signals, achieving closed-loop vehicle control.
\begin{equation}
{ Plan }_{ {path }}=\!\!\bigcup_{\tiny j=1}^{\tiny {N_d}\times{\frac{N_p}{2}}}\!\!\left\{\! \frac{ { point }_{ {l }}^{\mathrm{j}}+{ point }_{ {r }}^{\mathrm{j}}}{2} \bigg| { plan }_{l}^j,{ plan }_{r}^j=1\!\!\right\}\!.\!
\end{equation}

\subsection{Loss Function}
\label{loss}
\systemname predicts coarse priors and executes perception and planning, including: $L_{{roi}}$ for coarse lane prior prediction, $L_{{edg}}$ for double-edge 3D regression, $L_{{int}}$ and $L_{{dir}}$ for intersection and direction respectively, $L_{{occ}}$ and $L_{{plan}}$ for occupancy and planning respectively. Additionally, $L_{{spd}}$ and $L_{{sig}}$ are used for speed and traffic signals. Formulated as:
\begin{equation}
\begin{aligned}
Loss= \gamma L_{{roi}} + \delta L_{int} + \epsilon L_{dir} + \varepsilon L_{occ} +\\
\zeta L_{plan} + \eta L_{{edg}} + \theta L_{spd} + \iota L_{sig},
\end{aligned}
\end{equation}
where, in training, ${\gamma}$, ${\delta }$, ${\epsilon}$, ${\varepsilon}$, ${\zeta}$, ${\eta}$, ${\theta}$ and ${\iota}$ are set to 3:2:1:3:4:5:1:0.1. The coarse prior $L_{{roi}}$ is formulated as:
\begin{equation}
L_{{roi}}=\frac{1}{N_{gt}} \sum_{\tiny {i=0}}^{\tiny {N_{gt}-1}}\sum_{\tiny {j=0}}^{\tiny {{\frac{N_p}{2}}-1}} \left \{ \left | y_{ij}^{{l}}-\hat{y}_{ij}^{{l}} \right | + \left | y_{ij}^{{r}} - \hat{y}_{ij}^{{r}} \right | \right \}.
\end{equation}
For perception and planning, $L_{{edg}}$ utilizes the Manhattan distance for point regression, $L_{{int}}$, $L_{{dir}}$, and $L_{{occ}}$ employ Focal Loss~\cite{tool-focal}, while $L_{{spd}}$ uses SmoothL1Loss\cite{tool-fastrcnn-smooth-l1}, $L_{{sig}}$ is Cross-Entropy Loss, and $L_{{plan}}$ is formulated as: 
\begin{equation}
L_{plan}\!\!=\!\!\!\!\!\!\sum_{\tiny {i=0}}^{\tiny {N_{gt}-1}}\! \sum_{\tiny {j=0}}^{\tiny {{\frac{N_p}{2}}-1}}\! \!\!\left \{\!\! \frac{{\left ( \!\rho\!\cdot \!\left (\!1\!\!-\!e^{\tiny{\!-CE(y_{ij}},\hat{y}_{ij})}\!   \right ) \!\right )^{\tiny {2}}\! \!\!\cdot\! CE(y_{ij},\hat{y}_{ij})}}{D_{p2t}} \!\!\right \}\!\!,\!
\end{equation}
where $D_{{p2t}}$ represents the distance from an edge point in double-edge to the target point, weighting to emphasize planning features near targets, and $\rho$ is to 0.25. 

%% file: 5-experiments.tex
\section{EXPERIMENTS}
\begin{table}[t]
\centering
\caption{Performance Comparison on Carla Town05 Long.}
\setlength\tabcolsep{8.5pt}
    \begin{tabular}{l|c|ccc}
    \toprule
        \textbf{Method (Year)}& \makecell{\textbf{Modality}}& \makecell{\textbf{DS} \ $\uparrow$}& \makecell{\textbf{RC} \ $\uparrow$}  & \makecell{\textbf{IS} \ $\uparrow$} \\
        \midrule
     CILRS'19\cite{e2e-cilrs}& C& 7.8& 10.3& 0.75\\
     LBC'20\cite{e2e-lbc}& C& 12.3& 31.9& 0.66\\
     NEAT'21\cite{m2m-neat}& C& 37.7&62.1& -\\
     Roach'21\cite{m2m-roach}& C& 41.6& 96.4&0.43\\
     WOR'21\cite{m2m-wor}& C& 44.8& 82.4& -\\
     ST-P3'22\cite{m2m-stp3}& C& 11.5& 83.2& -\\
     Transfuser'22\cite{m2m-transfuser}& C+L& 31.0& 47.5& 0.77\\
     Interfuser'22\cite{m2m-interfuser}& C+L& 68.3& 95.0& -\\
     VAD'23\cite{m2m-vad}& C& 30.3&75.2& -\\
     ThinkTwice'23\cite{multi-modal-thinktwice}& C+L& 70.9& 95.5&0.75\\
     DriveAdapter'23\cite{multi-modal-driveadapter} & C+L& 71.9& 97.3&0.74\\
     DriveMLM'23\cite{multi-modal-drivemlm} & C+L& \underline{76.1}& \underline{98.1}&\underline{0.78}\\
     ReasonNet'23\cite{multi-modal-shao2023reasonnet}& C+L& 73.2& 95.9&0.76\\
         \midrule
     \textbf{\systemname \ ($ours$)}& C+L& \textbf{91.1}& \textbf{99.6}&\textbf{0.92}\\
    \bottomrule
    \end{tabular}
\label{tab:carla-t5-long}
\vspace{-10pt}
\end{table}
\subsection{Dataset and Metrics}
Using autonomous driving environment Carla\cite{carla}, we gather 126K frames from diverse scenarios across 8 maps and 13 weathers. The data are collected at 2Hz with vehicles equipped with four cameras, a LiDAR, an IMU, and a GPS. And, we annotate 3D edge points with attributes including intersection, direction, occupancy, and planning.
Driving Score (DS), Route Completion (RC), and Infraction Score (IS) as key metrics evaluating performance, where higher scores reflect better progress, safety, and adherence to rules, respectively.

\subsection{Performance on Carla Benchmark}
In this section, we employ the Town05 Long and Town05 Short benchmarks for closed-loop evaluation to demonstrate how \systemname leverages the lane-level Camera-LiDAR fusion method to achieve superior planning performance while maintaining high computational efficiency.
The Town05 Long benchmark comprises 10 routes, with each spanning approximately 1km. Conversely, the Town05 Short benchmark includes 32 routes, each measuring 70m. These benchmarks collectively assess the model's overall performance, demonstrating state-of-the-art capabilities in end-to-end planning.

\begin{table}[t]
\centering
\caption{Performance Comparison on Carla Town05 Short.}
\setlength\tabcolsep{14pt}
    \begin{tabular}{l|c|cc}
    \toprule
        \textbf{Method (Year)}& \makecell{\textbf{Modality}}& \makecell{\textbf{DS} \ $\uparrow$}& \makecell{\textbf{RC} \ $\uparrow$} \\
        \midrule
        CILRS'19\cite{e2e-cilrs}& C& 7.5& 13.4\\
        LBC'20\cite{e2e-lbc}& C& 31.0& 55.0\\
     NEAT'21\cite{m2m-neat}& C& 58.7& 77.3\\
     Roach'21\cite{m2m-roach}& C& 65.3& 88.2\\
     WOR'21\cite{m2m-wor}& C& 67.8& 87.5\\
     Transfuser'22\cite{m2m-transfuser}& C+L& 54.5& 78.4\\
     ST-P3'22\cite{m2m-stp3}& C& 55.1& 86.7\\
     Interfuser'22\cite{m2m-interfuser}& C+L& 95.0& 95.2\\
     VAD'23\cite{m2m-vad}& C& 64.3& 87.3\\
    ReasonNet'23\cite{multi-modal-shao2023reasonnet}& C+L& \underline{95.7}& \underline{96.2}\\
         \midrule
     \textbf{\systemname \ ($ours$)}& C+L& \textbf{96.7}& \textbf{98.3}\\
    \bottomrule
    \end{tabular}
\label{tab:carla-t5-short}
\vspace{-10pt}
\end{table}

\noindent\textbf{Driving Score.}
The driving score (DS) comprehensively assesses autonomous driving systems performance, combining route completion and infraction scores. In the Town05 Long and Short benchmarks, \systemname outperforms all other algorithms, including both camera-only and camera-LiDAR fusion methods, by achieving the highest driving score. Specifically, in the Town05 long benchmarks, our system achieves a driving score improvement of $15.0\%$ (Long) and $1.0\%$ (Short), as detailed in Table \ref{tab:carla-t5-long} and \ref{tab:carla-t5-short}. These results underscore the effectiveness of \systemname in enhancing safety and traffic regulation compliance.

\noindent\textbf{Route Completion.}
The route completion (RC) is a critical metric for assessing an autonomous driving system's success in completing predetermined routes throughout the evaluation process, reflecting the system's ability to plan routes and accurately understand target points. In the Carla Town05 Long and Short benchmark, compared to other algorithms, our method shows a higher route completion rate in benchmark tests that included a diverse range of traffic scenarios. Specifically, our method achieves improvements of $1.5\%$ and $2.1\%$, respectively, as shown in Table \ref{tab:carla-t5-long} and \ref{tab:carla-t5-short}. These results highlight the advanced capabilities of our method in route planning, as well as its robustness in adapting to various road conditions.

\noindent\textbf{Infraction Score.}
The infraction score (IS) is a comprehensive metric used to evaluate systems performance, including avoiding collisions, adhering to traffic rules, and handling complex situations. 
\systemname exhibits significant performance in Town05 Long, achieving a $14.0\%$ improvement in infraction score (IS) as detailed in Table \ref{tab:carla-t5-long}, surpassing other algorithms.

\begin{table}[t]
	\centering
    \caption{ Comparative Analysis of Efficiency.}
    \setlength\tabcolsep{9.5pt}
	\begin{tabular}{c|c|c|c}
        \toprule
        \textbf{Method (Year)}& \textbf{Modality}& \textbf{Latency (ms)  $\downarrow$} & \textbf{FPS  $\uparrow$}  \\
        \midrule
        NEAT'21\cite{m2m-neat} & C &  85.08 & 11.75  \\
        ST-P3'22\cite{m2m-stp3} & C & 476.74 & 2.1  \\
        TransFuser'22\cite{m2m-transfuser} & C+L & 171.93 & 5.82  \\
        VAD'23\cite{m2m-vad} & C&  59.50 & 16.81  \\
        ThinkTwice'23\cite{m2m-transfuser} & C+L & 262.77 & 3.81  \\
        PHP'24\cite{php} & C& 44.30 & 22.57  \\
        \midrule
        \textbf{\systemname (ours)} & C+L& 51.90 & 19.27 \\
        \bottomrule
	\end{tabular}
      \vspace{-10pt}
    \label{tab:efficiency}
\end{table}

\begin{table}[t]
	\centering
    \caption{Comparative Analysis of LiDAR Feature Counts.}
    \setlength\tabcolsep{10.2pt}
	\begin{tabular}{c|c|c}
        \toprule
        \textbf{Method} & \textbf{Resolution} & \textbf{LiDAR Features}  \\
        \midrule
        LiDAR Voxel & [0.5, 0.5, 0.5] & 4364  \\
        LiDAR Pillar &  [0.5, 0.5, 8] & 3446  \\
        \midrule
        Lane-Level LiDAR Pillar &  [0.5, 0.5, 8] & \textbf{600}  \\
        \bottomrule
	\end{tabular}
      \vspace{-10pt}
    \label{tab:lane-sample}
\end{table}

\begin{figure}[t]
  \centering
  \includegraphics[width=0.98\linewidth]{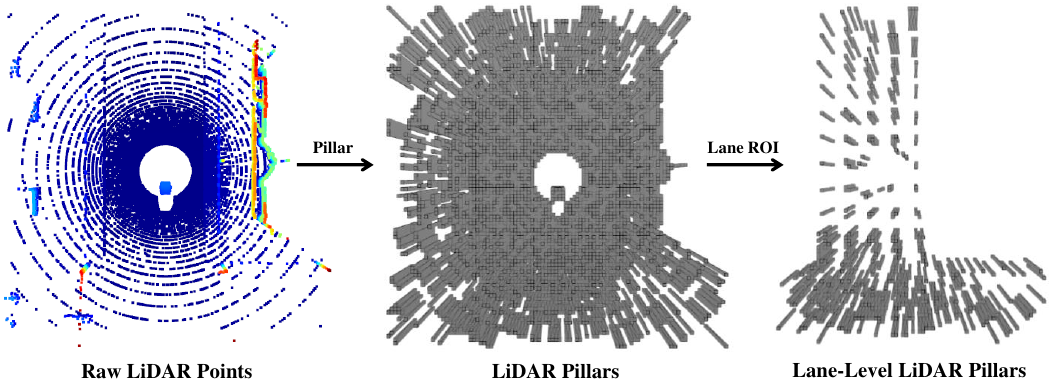}
  \caption{Visualizing the Transformation from LiDAR to Lane-Level Pillars}
  \label{fig:lane-level-lidar-pillar}
\end{figure}

\noindent\textbf{Computational Efficiency.}
Fusing LiDAR and camera data can enhance system performance. However, improper integration of LiDAR data may reduced operational efficiency. Our method employs a lane-level fusion algorithm that extracts and integrates critical depth features required by the image branch, thereby enabling the system to process only essential depth data. This approach not only reduces computational redundancy in the LiDAR branch but also minimizes the cost of multi-modal data fusion, achieving efficient data integration.
As shown in Table~\ref{tab:efficiency}, our algorithm achieves a computational speedup of 19.27 FPS over other camera-LiDAR fusion methods, with the LiDAR branch's introduction adding only 7.6 ms overhead. Furthermore, Figure \ref{fig:lane-level-lidar-pillar} visualizes the lane-level LiDAR pillar results, showing that depth data from only critical lane areas are extracted. Table \ref{tab:lane-sample} quantitatively describes the amount of features obtained by different LiDAR processing methods. After processing, the lane-level LiDAR pillars reduce the data volume by 7.27 times compared to Voxel and 5.74 times compared to Pillar methods. This results in higher processing efficiency for the LiDAR branch.

\subsection{Qualitative Results}
\begin{figure}[tp]
  \centering
	\centering
	\begin{minipage}{0.49\linewidth}
		\centering
		\includegraphics[width=\linewidth]{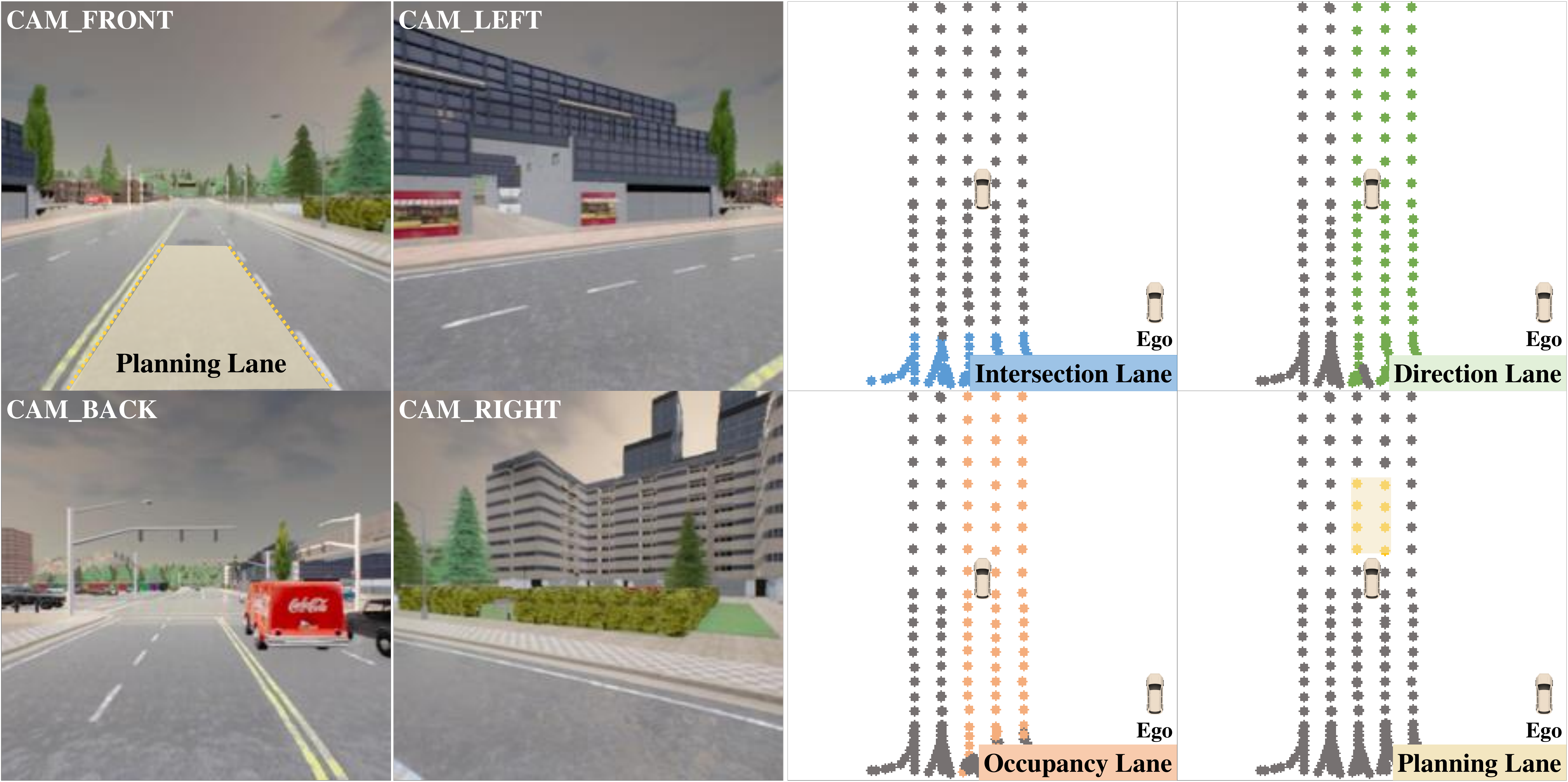}
	\end{minipage} 
	\begin{minipage}{0.49\linewidth}
		\centering
		\includegraphics[width=\linewidth]{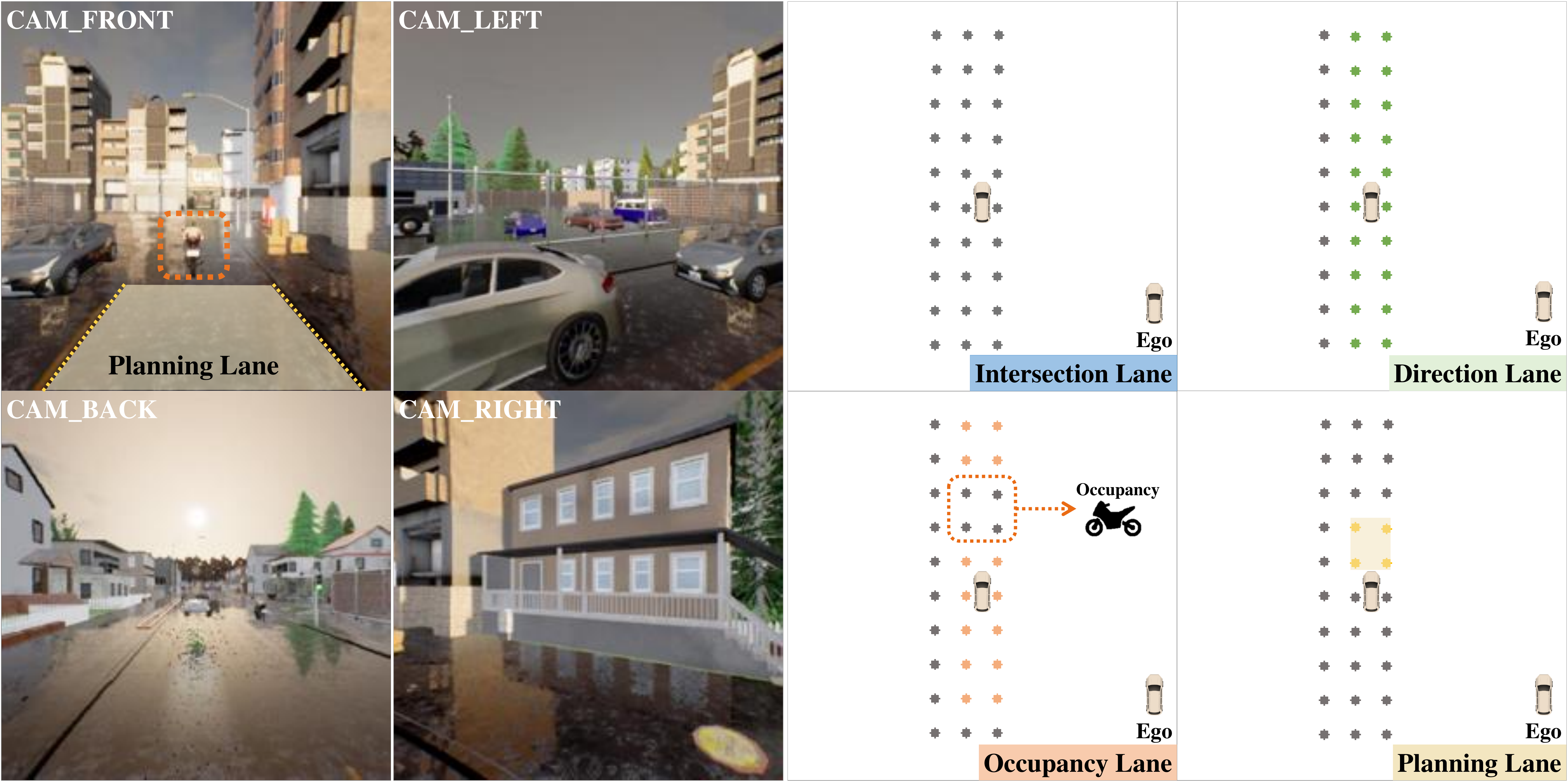}
	\end{minipage}
	\begin{minipage}{0.49\linewidth}
		\centering
		\includegraphics[width=\linewidth]{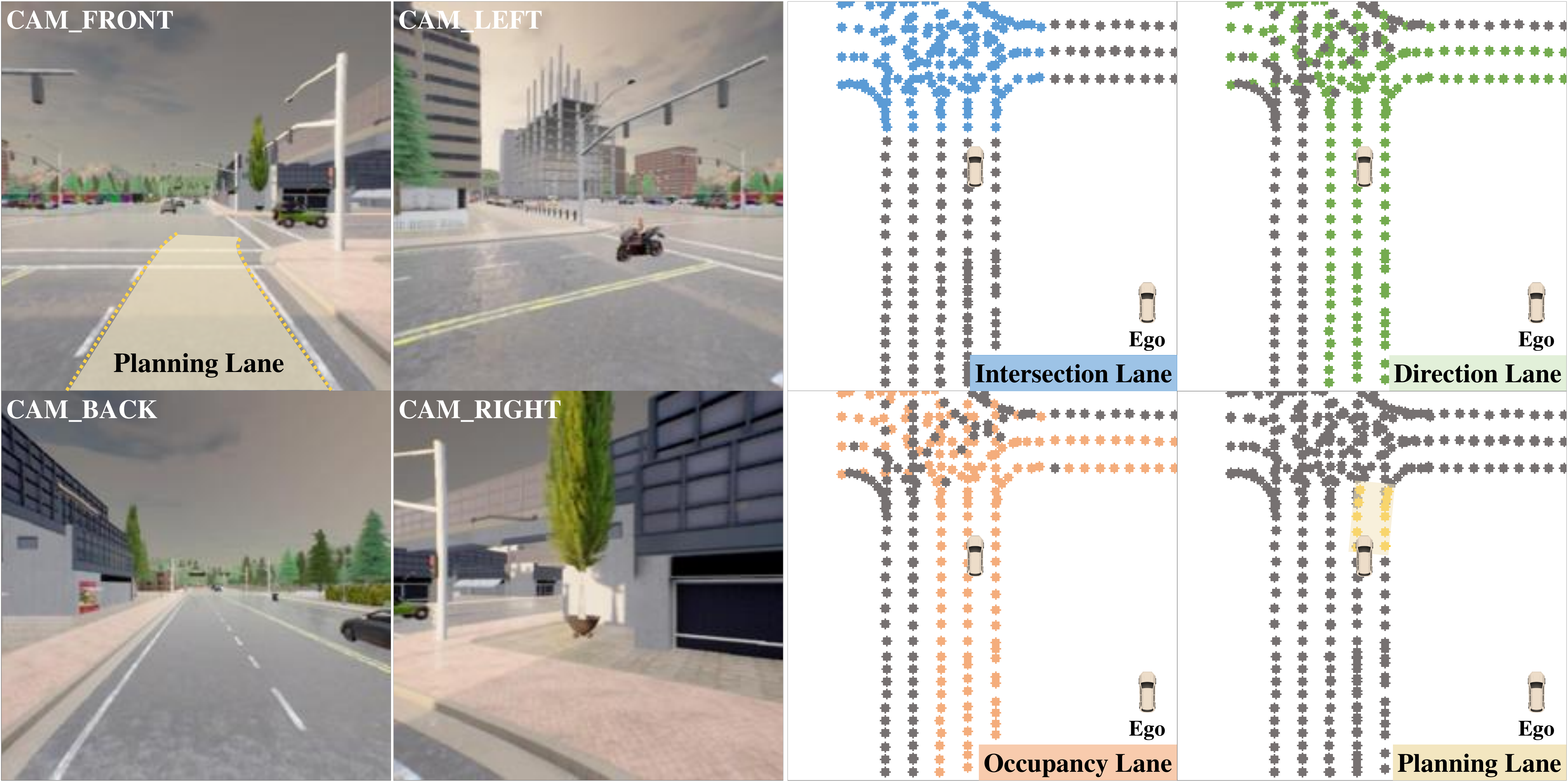}
	\end{minipage}
	\begin{minipage}{0.49\linewidth}
		\centering
		\includegraphics[width=\linewidth]{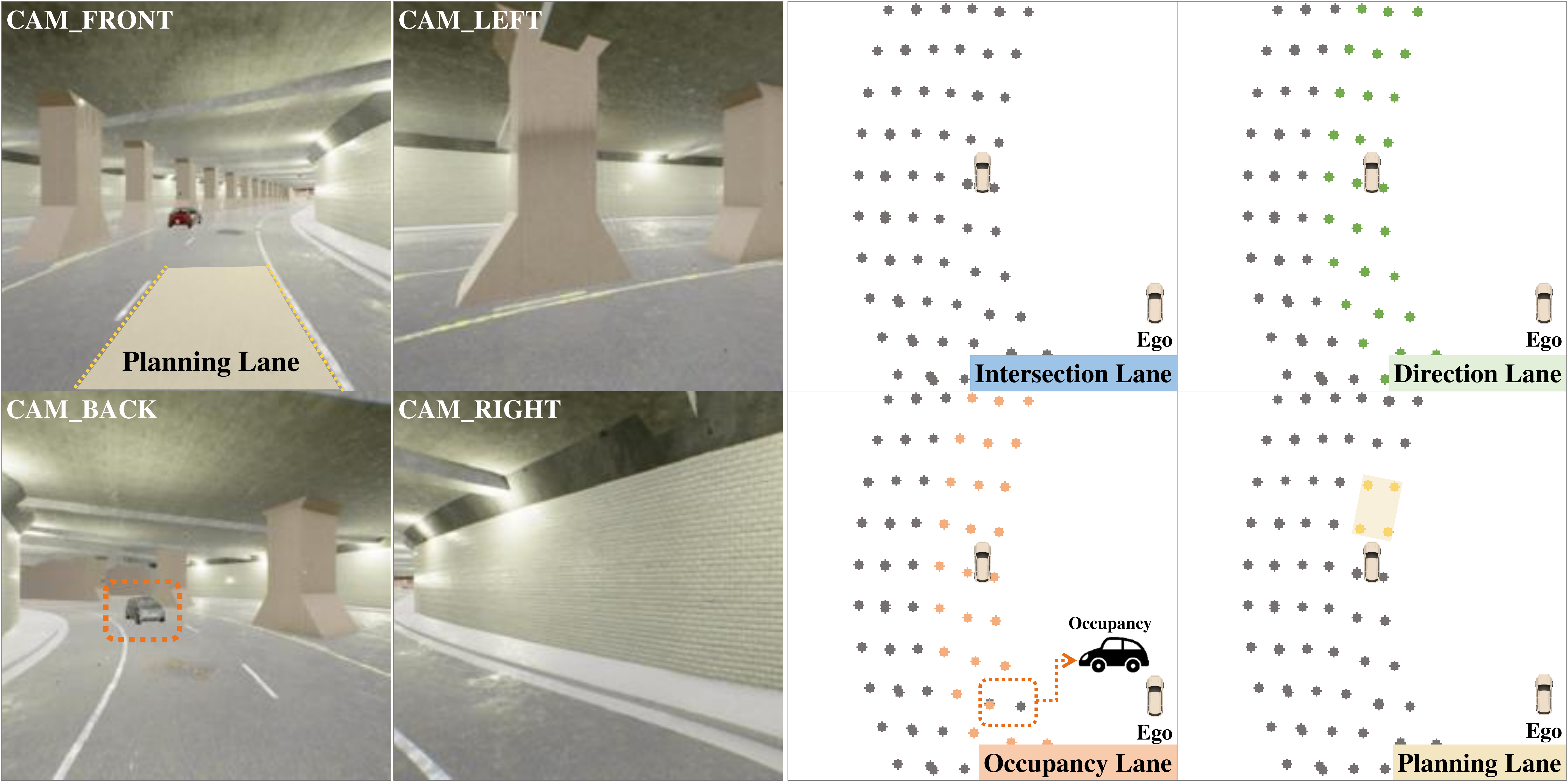}
	\end{minipage}
   \caption{Visualization of Lane-Level Perception and Planning in \systemname. 
   }
  \label{fig:exp_vis}
  \vspace{-10pt}
\end{figure}

In Figure \ref{fig:exp_vis}, we illustrate our system's ability to hierarchically perceive the traffic environment and seamlessly integrate planning tasks at the lane level. The visualization includes intersection lanes marked in blue, direction lanes that indicate roads complying with traffic regulations marked in green, and occupancy lanes for roads unoccupied by traffic agent and adhering to direction marked in orange. The planning lane, highlighted in yellow, signifies the optimally chosen lane that ensures safety and leads to the target point.

\subsection{Ablation Studies}

\begin{table}[t]
\centering
\caption{Ablation study on Carla Town05 Long. ``CLM" denotes coarse lane detection module, ``CQF" denotes lane-level cross-modal query integration and feature enhancement, and ``LFE" denotes lane-level LiDAR feature extraction.}
\setlength\tabcolsep{7pt}
    \centering
        \begin{tabular}{c|ccc|ccc} 
        \toprule
        \textbf{ID}&  \textbf{CLM}&  \textbf{CQF}&  \textbf{LFE}&\multicolumn{3}{c}{\textbf{Carla Town05 Long}}\\
        \midrule
        \textbf{--}&Sec\ \ref{method:img-guide-lane-prioritization}&Sec\ \ref{method:lane-level-cross-modal}&Sec\ \ref{method:lane-level-lidar-featrue}&  \textbf{DS$\uparrow$}&  \textbf{RC$\uparrow$}&  \textbf{IS$\uparrow$}\\ 
        \midrule
        A&\tiny{\color{red}\XSolidBrush}&\tiny{\color{green}\CheckmarkBold}&\tiny{\color{green}\CheckmarkBold}& 17.4& 54.4& 0.46\\ 
        B&\tiny{\color{green}\CheckmarkBold}&\tiny{\color{red}\XSolidBrush}&\tiny{\color{green}\CheckmarkBold}& 50.2& 87.0& 0.57\\ 
        C&\tiny{\color{green}\CheckmarkBold}&\tiny{\color{green}\CheckmarkBold}&\tiny{\color{red}\XSolidBrush}& 78.3& 96.2& 0.81\\ 
         \midrule
        \textbf{D}&\tiny{\color{green}\CheckmarkBold}&\tiny{\color{green}\CheckmarkBold}&\tiny{\color{green}\CheckmarkBold}&  \textbf{91.1}&  \textbf{99.6}&  \textbf{0.92}\\
         \bottomrule
        \end{tabular}
    \label{tab:ablation}
\end{table}

\noindent\textbf{Impact of Coarse Detection Lane Module.}
This module utilizes image features to provide coarse lane ROI priors to the LiDAR branch, enabling it to learn critical depth information. To validate the module's influence, we conducted an experiment (Experiment A) where the learnable coarse lane priors were replaced with random lane ROI priors.
Our findings indicate that random priors fail to direct the LiDAR branch towards the depth information of interest to the image branch, resulting in the fusion features lacking the desired depth cues. This deficiency leads to a degradation in performance. As shown in Table \ref{tab:ablation}, Experiment A's results demonstrate a decrease in driving score (DS), route completion (RC), and infraction score (IS) by 73.7\%, 45.2\%, and 46\%, respectively, compared to the original learnable module.
This experiment underscores the importance of accurate lane ROI priors in enhancing the depth perception capabilities of the LiDAR branch and the overall performance of the \systemname method.

\noindent\textbf{Impact of Lane-Level Cross-Modal Query Integration and Feature Enhancement. }
This module is hinged on an adaptive weighting fusion mechanism, which we demonstrate as crucial for effective cross-modal fusion. 
In Experiment B, we removed the weighted query integration, opting to use only the initial LiDAR query. We also replaced the weighted feature enhancement with an equal-weight fusion of dual-branch features, which neutralized the depth information prioritization.
This adjustment resulted in notable decreases in key metrics: driving score, route completion, and infraction score by 40.9\%, 12.6\%, and 35\%, respectively, as shown in Table \ref{tab:ablation}. The results affirm that adaptive weighting of lane priors is essential for the LiDAR branch to accurately extract and integrate depth features, thus enhancing the multi-modal system's fusion efficiency and overall performance.

\noindent\textbf{Impact of Lane-Level LiDAR Feature Extraction. }
This module enhances the \systemname by providing depth features complementary to the image branch. Removing this module reverts \systemname to a camera-only approach (Experiment C), which, as shown in Table \ref{tab:ablation}, suffers from a lack of depth information,  causing decreases in driving score, route completion, and infraction score by 12.8\%, 3.4\%, and 11\%, respectively. This highlights the module's critical role in leveraging the complementary strengths of multi-modal data.

%% file: 6-conclusions.tex
\section{CONCLUSIONS}
In this paper, we presented the end-to-end lane-level camera-LiDAR fusion planning (\systemname) method, which effectively balances performance with computational efficiency. \systemname enhances efficiency by integrating image-guided lane prioritization with sparse LiDAR sampling, leveraging lane priors to minimize computational redundancy and focus on lane-relevant data. Additionally, \systemname employs an efficient lane-level query integration and feature enhancement strategy, effectively merging semantic image information with LiDAR depth data to produce an comprehensive lane-level representation essential for planning.
Experiments on the Carla Benchmark validate \systemname's superior efficiency and performance, surpassing current algorithms. Our lane-level camera-LiDAR fusion strategy ensures operational excellence at 19.27 FPS, underscoring the benefits of our integrated approach.